\documentclass[journal]{IEEEtran}

\ifCLASSOPTIONcompsoc
  \usepackage[nocompress]{cite}
\else
  \usepackage{cite}
\fi

\ifCLASSINFOpdf
  \usepackage[pdftex]{graphicx}
  \graphicspath{{figs/}}
  \DeclareGraphicsExtensions{.pdf}
\else
\fi

\usepackage[cmex10]{amsmath}
\usepackage{amssymb}
\usepackage[caption=false,font=footnotesize]{subfig}
\usepackage{balance}

\newlength{\imagewidth}

\begin{document}

\title{Discovering Latent States for Model Learning: Applying Sensorimotor Contingencies Theory and Predictive Processing to Model Context}

\author{Nikolas~J.~Hemion
\IEEEcompsocitemizethanks{\IEEEcompsocthanksitem Nikolas J. Hemion is with the AI
Lab of SoftBank Robotics Europe (\mbox{Aldebaran}), 43 rue du Colonel Avia, 75015
Paris, France.\protect\\
E-mail: nhemion@aldebaran.com}
}

\maketitle

\begin{abstract}

Autonomous robots need to be able to adapt to unforeseen situations and to acquire new skills through trial and error. Reinforcement learning in principle offers a suitable methodological framework for this kind of autonomous learning. However current computational reinforcement learning agents mostly learn each individual skill entirely from scratch. How can we enable artificial agents, such as robots, to acquire some form of generic knowledge, which they could leverage for the learning of new skills? This paper argues that, like the brain, the cognitive system of artificial agents has to develop a world model to support adaptive behavior and learning.  Inspiration is taken from two recent developments in the cognitive science literature: predictive processing theories of cognition, and the sensorimotor contingencies theory of perception.  Based on these, a hypothesis is formulated about what the content of information might be that is encoded in an internal world model, and how an agent could autonomously acquire it. A computational model is described to formalize this hypothesis, and is evaluated in a series of simulation experiments.

\end{abstract}


\IEEEpeerreviewmaketitle

\section{Introduction}
\label{sec:introduction}

\IEEEPARstart{A}{utonomous} robots need to be able to adapt to unforeseen
situations and to acquire new skills through trial and error.  Reinforcement
learning in principle offers a suitable methodological framework for this kind
of autonomous learning \cite{sutton1998reinforcement}. However there is
undoubtedly still a significant difference between the learning performance of
current computational reinforcement learning agents and that of their
biological counterparts (humans and other animals): the latter can readily make
use of a huge amount of \emph{generic} knowledge, which they have accumulated
over the whole history of their past experience, allowing them to quickly come
up with strategies to solve novel tasks, whereas current computational
reinforcement leaning agents mostly learn each individual skill entirely from
scratch. How can we endow artificial agents, such as robots, with the
capability to acquire some form of generic knowledge, which they could leverage
for the learning of new skills?

This paper argues that the difference in learning performance is largely due to
the lack of a generic model-acquisition mechanism in artificial systems, and
offers a perspective on the topic, which is taking inspiration from recent
developments in the cognitive science literature. In particular, this work is
motivated by ``predictive processing (PP)'' theories of cognition
\cite{wolpert1998internal, friston2007free, clark2013whatever,
seth2014predictive}, according to which a fundamental principle behind the
functioning of the brain is the continuous minimization of prediction errors,
or equivalently ``free energy'' \cite{friston2007free}. As a consequence of
this drive, so it is argued by the theories, the brain acquires a hierarchical
model that estimates the latent causes of observable events. PP has been well
received by the recent cognitive science literature and related disciplines as
a candidate brain theory \cite{friston2010free}.  However, it describes
cognitive processing on a rather high level of abstraction and leaves most
``implementation details'' open for interpretation.  In particular, PP does not
provide a clear account for how the latent causes of observable events could be
learned by an agent without any form of prior knowledge, and to what extent an
agent should try to estimate latent causes -- an agent can certainly not learn
everything, but has to focus on what is relevant for its behavior.

It has recently been proposed \cite{seth2014predictive,
laflaquiere2015grounding, seth2015cybernetic} that PP lends itself naturally to
be combined with the so called ``sensorimotor contingencies theory (SMCT)'' of
perception \cite{o2001sensorimotor}, which might shed some light on aspects of
PP that are underspecified. In tandem, PP and SMCT can offer a more concrete
hypothesis about what could be the basic units of knowledge in the brain:
predictive models that capture information about how certain sensory input
signals change in a particular way, in response to own actions.  Importantly,
these predictive models are \emph{generic}, as their acquisition and
functioning does not depend on a certain task or modality. Furthermore they are
\emph{contextual}: their outputs are only correct in certain contexts (or
``contingencies'', in the language of the SMCT). Said otherwise, they only
predict well as long as some external situation is in place.

A simple example is that of object perception: A good predictive model
capturing information about what an object looks like could produce accurate
predictions for the visual input when the agent is looking at the object, and
how this input will change when the object is moved. But the predictions of
this model would only be correct as long as the object is actually present.
When the object is replaced with a different one, the model would of course no
longer be able to make accurate predictions. The model itself can thus be seen
as an active component in the cognitive system of the agent It corresponds to a
hypothesis of the cognitive system about an environmental circumstance: it
captures the hypothesis that a certain object is present, or in the general
case, the hypothesis of what could be called a ``sensorimotor context'' (cf.
also the model presented in \cite{haruno2001mosaic}, one of the early
precursors of the predictive processing hypothesis).

In this paper, these ideas will be further developed, and importantly, it will
be explored how a naive agent with no prior experience can autonomously learn
the hypothesized basic units of knowledge. A computational model will be
presented to demonstrate and evaluate the ideas developed in this paper. It
treats the experiences of an agent as transitions in a graph of sensorimotor
states. Candidate models are densely connected subgraphs within this graph. It
will be explained, how this computational model relates to the concepts behind
PP and the SMCT.  The computational model is evaluated in a simplistic
simulation, which was specifically designed to highlight several important
aspects of the learning problem that the agent is facing. In particular, the
agent has to deal with ambiguity in sensorimotor states, a difficulty that will
be described in detail along with a solution.

\section{Related Work and Scientific Background}
\label{sec:related_work}

\subsection{Computational Reinforcement Learning}

In the last two decades, computational reinforcement learning has stepped out
of the realm of simple simulated grid worlds, and has entered the real world
through physical robots: many algorithms that are based on the core principles
of reinforcement learning (i.e. iteratively improving a behavior through
trial-and-error learning, maximizing a reward signal) have been proposed in the
robotics literature, and have been successfully applied to robot skill learning
in a range of experiments \cite{kober2013reinforcement}. However, a key
requirement that enables the success of any of these algorithms is the
definition of a compact state-space: a re-description of the \emph{important
aspects} of the actual physical problem, in a low-dimensional and often highly
task-specific abstract vector space. In a way it can be said that through this
definition of a state-space, the human designer already endows the robot with
large amounts of her or his generic knowledge (and with it, parts of the
solution), before the actual learning has even begun. Naturally, this
significantly simplifies the learning problem. In fact, it has recently been
shown \cite{stulp2013robot} that many of the successes that we have seen are
not due to improvements in learning algorithms, but much rather due to the way
that researchers have learned to simplify the learning problem so much
\cite{ijspeert2002movement} that even simple black-box optimization algorithms
are able to successfully solve the problem, and in some cases even outperform
highly sophisticated learning algorithms. While these approaches have led to
robots demonstrating impressive performance in terms of dexterity and
precision, the fact that they depend on the manual definition of a state-space
(i.e. feature representation of the sensorimotor space) prohibits their
application as generic learning methods, which would be required for truly
autonomous robot learning.

To overcome the necessity for manually designed feature representations, one
might argue that unsupervised learning and dimensionality reduction methods
could be employed, as it might be possible to discover suitable and generic
features in a data-driven way. The upsurge and success stories of deep learning
methods in the computer vision and machine learning communities
\cite{bengio2009learning} could be taken as arguments in favor of such a view.
And indeed, it has recently been demonstrated that it is possible to not only
learn a policy, but at the same time also a suitable abstraction from raw
sensory inputs, thus tackling and solving the learning problem ``end-to-end''
\cite{levine2015end, mnih2013playing}. This is achieved by extending the
reinforcement learning system with a deep learning component. Through
optimization, these systems are capable to extract a suitable feature
representation that is relevant to the task at hand, thus learning a behavior
while only minimally relying on human task-knowledge.  This is a major step
forward from solutions that rely on hand-specified feature representations.
But nonetheless, the same fundamental problem remains: the information that is
encoded in the system after optimization is highly task-specific, and as a
consequence, the agent still has to learn every new task almost entirely from
scratch.

Another perspective that can be taken on the topic is that of ``transfer
learning'', which is concerned with the problem of how to transfer information
from one domain (the source domain) to another (the target domain), see
\cite{taylor2009transfer} for an overview. Ultimately, if it were possible to
transfer knowledge in a \emph{generic} way between domains and behaviors, this
could be seen as a solution to the problem of how to endow an agent with
generic knowledge. However, the majority of approaches in the body of
literature on transfer learning assume that the state space of source and
target domain are identical and manually defined in advance. Thus, these
approaches do not directly contribute to answering the question of how an agent
could acquire generic knowledge. Other works assume different state spaces in
source and target domain but either require a manually defined transfer mapping
between these (still predefined) spaces, or some other form of task-specific
knowledge. There are a few notable exceptions \cite{ferguson2006proto,
taylor2008transferring}, but they are concerned with the question of how to
transfer information between two tasks that are in some way equivalent, which
is not the case in general.

\subsection{PP and the SMCT}

What is still missing in any of the above described systems, so it seems, is a
mechanism to acquire \emph{generic} knowledge, which could be leveraged for the
learning of new skills. In the following, relevant aspects of PP and SMCT will
briefly be summarized, with the aim to motivate a choice of learning algorithm
for a cognitive agent.

\label{sec:predictive_processing}
\label{sec:pp}
\label{sec:smct}

Predictive processing, in form of the ``free energy'' principle, has been
proposed as a candidate for a unifying brain theory \cite{friston2007free,
friston2010free} and has received much attention in the cognitive science
literature in recent years (see also \cite{seth2015cybernetic} for a more
thorough discussion of the historical roots of the theory). The overall idea is
that through evolution, brains have developed into a means of adaptive agents
to minimize their likelihood to encounter harmful situations.  To achieve this,
the brain requires the capacity to discover the causes of sensory observations,
in order to allow the agent to generate adaptive responses. However, it does
not have direct access to information about the true causes, but has to rely
solely on the sensory information that it receives. According to PP, the brain
deals with this problem by constructing a hierarchical generative model and
performing probabilistic inference about what are the causes of sensory
observations. It thus builds up an internal model of latent environmental
causes, mirroring the informational structure of the external world. The brain
then operates by producing top-down predictions of future sensory inputs, and
comparing them with actually observed inputs. This way, it generates prediction
errors, which are used to update the hierarchical predictive model to better
match what is actually observed. This view on the nature of perception as a
process largely driven by top-down signals is in stark contrast with
traditional views of perception as a bottom-up information processing
``pipeline''.

While indirect empirical support for this interpretation of brain function is
accumulating \cite{seth2014predictive}, the theory is lacking a clear
description of how the internal model is constructed, and of the nature of the
information that is encoded in the hierarchical model.

The sensorimotor contingencies theory \cite{o2001sensorimotor} is another
prominent theory of perception in the recent cognitive science literature, and
also constitutes a radical break with traditional views of perception.
According to the theory, perception corresponds to the ``mastery of
sensorimotor contingencies'' (dynamic interactions of the agent with its
environment). An often recited example that is meant to illustrate this rather
vague description is that of color perception: perceiving something as
\emph{red}, so the argument, does not correspond to the circumstance that light
of a certain wave-length is impacting the cones on the retina. Actually, light
of the same wave-length can under certain conditions be reflected off of
different surfaces that are actually perceived as not having the same color.
What actually amounts to the perception of the color red, according to the
theory, is the knowledge of how the incoming stimuli \emph{change} when the
agent interacts with the environment (for example by moving the red surface and
thus dynamically changing the reflection of ambient light).

The theory can be interpreted in the following way, which helps to formulate
the concepts the theory provides in a robotics or machine learning context: An
important consequence of the theory is that perception (for example of a
certain color) is not a property of the stimulus, nor is it a property of the
sensor. Instead, perception is said to be a property of an interaction of the
agent with its environment, or rather of the ``mastery'' of this interaction,
which is a sensorimotor contingency. In fact, substantially different sensors
and motors could be employed by an agent to engage in the \emph{same}
sensorimotor contingency, and would thus result in the same perception
\cite{o2001sensorimotor, laflaquiere2013learning}.  Through its ``mastery'' of
contingencies, an agent has the means to recognize external reality. It can
engage in the interaction with the environment that is specific to a
contingency, and as long as this interaction succeeds, the contingency is
perceived by the agent. The agent's knowledge of contingencies can thus be
characterized as strategies, or policies, to test specific hypotheses about its
environment. By following such a strategy, it can verify the hypothesis that
the contingency is indeed in place, or falsify it by noticing that the
interaction fails.

As compared to PP, the SMCT lacks a clear description of the underlying
processes on the level of neuron activations or in the form of a mathematical
formalization. In particular, it is not clearly specified what ``mastery''
means in concrete terms. On the contrary, SMCT strongly highlights that
perception is intimately coupled with action (hence the name
``sensori\emph{motor} contingencies theory''). Furthermore, in contrast to PP,
it makes a distinction about what \emph{is} perception and what \emph{is not}:
only contingencies that the agent can itself influence through its own actions
are relevant for it, whereas arbitrary but recurring patterns of signals, which
might occur naturally but without any possibility for the agent to influence,
are meaningless for the agent and are thus not perceived.

\section{A Hierarchical Predictive Model of Sensorimotor Contingencies as Representational Framework}

Both PP and the SMCT lack a level of detail that is sufficient for a
comprehensive computational implementation. While there exist a number of
computational models that serve to demonstrate important properties of the two
theories (e.g. \cite{friston2012perceptions, philipona2006color,
laflaquiere2015learning}), none of them addresses the fundamental question of
how a naive agent can autonomously develop the internal structure that is
postulated by the respective theory. But taking the two theories together, a
picture of a cortical system emerges that moves us closer towards a holistic
understanding, as will now be outlined. The described view will also allow us
to develop a computational model, which will be described in
Section~\ref{sec:computational_model}.

By combining the views of PP and the SMCT, we can hypothesize that the
cognitive system of an agent should constitute of a hierarchical predictive
model that is employed to predict future sensory signals. In this model,
hierarchically higher levels estimate latent causes for information that is
encoded on respective lower levels, or said otherwise, hypotheses about
lower-level contingencies.

\begin{figure}
\setlength{\imagewidth}{0.46\linewidth}
\begin{center}
    \hfill
    \subfloat[]{
        \includegraphics[width=\imagewidth]{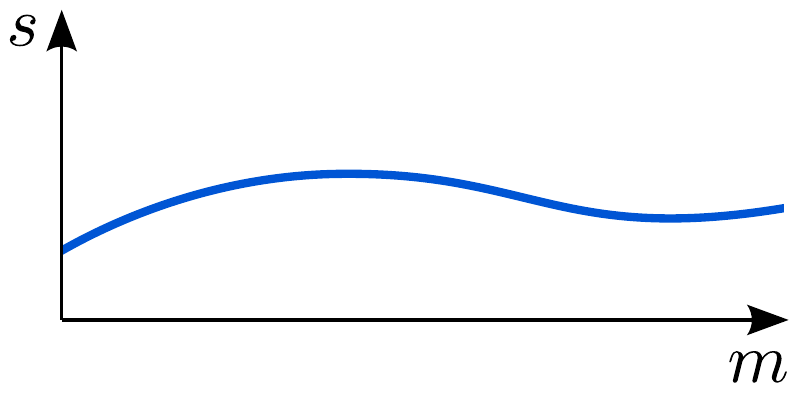}
        \label{subfig:ex_sm_interaction_1a}
    }
    \hfill
    \subfloat[]{
        \includegraphics[width=\imagewidth]{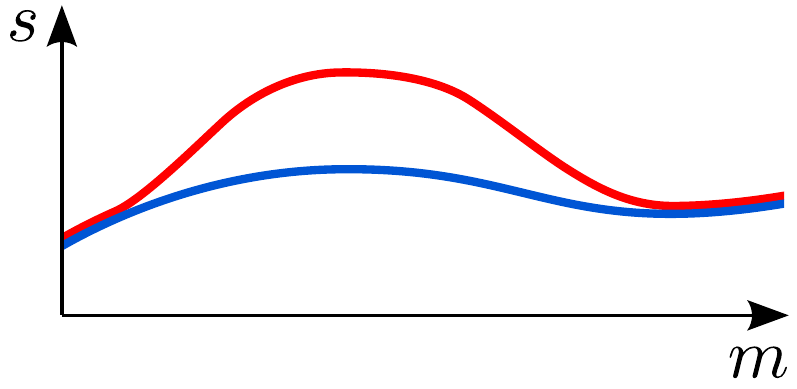}
        \label{subfig:ex_sm_interaction_1b}
    }
    \hfill\phantom{.}
\end{center}

\caption{Abstract examples of sensorimotor interactions, see text for description.}

\label{fig:ex_sm_interaction_1}
\end{figure}

To make this a bit clearer, consider the abstract example of a sensorimotor
interaction between an agent and its environment that is depicted in
Figure~\ref{subfig:ex_sm_interaction_1a}, where the horizontal and vertical
axes correspond to the agent's motor and sensor spaces, respectively. When the
agent maneuvers through its motor space (i.e. it translates its motor state
left and right along the horizontal axis), it observes a stable sensory
feedback pattern, which can be plotted as a curve. In this very simple example,
the agent can easily learn a predictive model to capture the laws of its
sensorimotor interaction.  There is no reason for the agent to assume that the
environment has a state, which is different from the agent's own internal
state. Indeed, the agent does not even have any reason to assume that there is
any such thing as an environment. From the agent's point of view, the whole
world seems perfectly deterministic and completely controllable. Furthermore,
there is no point for this agent to maintain a memory of any sort, such as an
\emph{internal} state (as opposed to its \emph{physical} state, i.e. the one of
its body). The agent could perfectly control its sensorimotor apparatus without
having to know anything about the past.

Now consider the case of another agent, whose sensorimotor interaction with the
environment is depicted in Figure~\ref{subfig:ex_sm_interaction_1b}. It has
identical motors and sensors as the agent from the first example, but its
situation is substantially different: for some of the configurations in its
motor space, it always observes a stable sensory feedback, just like the first
agent, but for a subpart of its motor space it observes two different sensory
feedbacks for identical motor configurations. This second agent faces the
problem that was outlined above in the description of the PP account for
cognition (see Section~\ref{sec:pp}): the sensory signals that it observes
depend on factors that are external to the agent itself (changes happening in
the agent's environment), and thus are not controllable by the agent.
Furthermore, the agent has no means to directly measure the external factors.
Were it to make memoryless predictions of sensory observations (like the first
agent), it would perform at chance level at most.  Its only way to improve its
capability to make accurate predictions is by maintaining an internal
representation that somehow estimates the external causes of sensory
observations.

Many formalisms have been introduced in the literature to address this problem.
For example, one common way (among many others) to include an internal state in
a predictive model is to use a recurrent neural network, which incorporates
temporal information into its predictions. The point here however is not simply
that this is a problem that needs to be addressed in some way. Instead, this
work looks at the bigger picture: what is the nature of the representational
system of the brain, and how could it be implemented in an artificial agent?
Without a clear connection to a comprehensive theory of brain organization, it
is difficult to see how a computational model might be extended to a complete
cognitive system.

PP and the SMCT offer such a theory of brain organization, to a sufficient
extent. Following the above interpretation of the theories, we can postulate
that the internal representation of external causes of sensorimotor
observations is in the form of internal states, which encode hypotheses about
how the flow of sensorimotor signals will unfold. Concretely, we can imagine
the agent to store two internal models, each of which estimates one of the two
curves that are plotted in Figure~\ref{subfig:ex_sm_interaction_1b}. By
assigning an activation level to each one of these internal models, the agent
becomes equipped with a simple form of memory and the ability to test its
hypotheses about external causes: the agent can make predictions about how its
sensory input will change in response to changing its motor state. Sensory
observations will lie on either one of the two curves, which will lead to the
corresponding internal model to be selected. In this view, the agent, after
learning, would not simply make different sensory observations, but it would
\emph{perceive} two distinct \emph{sensorimotor contexts}, or
\emph{contingencies}.

It was already demonstrated in the literature that such a selection mechanism
can be implemented (e.g. \cite{haruno2001mosaic}). This paper addresses the
problem of how an agent can autonomously discover the external state for the
training of appropriate internal models. For the purpose of demonstrating how
this can be achieved, the next section will introduce a mathematical
formalization of the problem description.

\section{Mathematical Formalization}

We can formally describe the sensorimotor experience of an agent in the
following way: as it explores its sensorimotor space, for example by following
an exploratory control policy (such as \cite{oudeyer2007intrinsic}), the agent
observes sequences of sensorimotor states $\mathbf{x}_i$. Here, $\mathcal{X} =
\{ \mathbf{x}_i \}$ is the set of possible sensorimotor states and corresponds
to extero- as well as interoceptive signals, which also includes efferent
copies of motor signals. In the examples described above, this set of
sensorimotor states is that of all possible pairs $(m_i, s_i) = \mathbf{x}_i$
in the two-dimensional plane that is depicted in
Figure~\ref{fig:ex_sm_interaction_1}.

We can describe the experience of the agent, as it follows an exploration
policy $\pi$, as a stochastic process:
\begin{equation}
X^{\pi} = \left\{ X_t = \mathbf{x}_i~:~t = 1, \dots, n \right\}.
\label{eq:stochastic_process}
\end{equation}
If we further assume that the set $\mathcal{X}$ of sensorimotor states is
countable (it will later be shown how to relax this assumption to deal with
continuous sensorimotor spaces), we can model the agent's sensorimotor
exploration as a discrete-time Markov chain $X_1, X_2, \dots, X_n$.

In this formulation, the task of the agent to minimize its error in predicting
future states amounts to estimating the probability
\begin{equation}
    \Pr( X_{t+1} = \mathbf{x}_j ~ | ~ X_t = \mathbf{x}_i, \pi, E_t = \mathbf{e}_k ),
    \label{eq:state_transition_probability}
\end{equation}
which describes the probability of observing a transition from one sensorimotor
state, $\mathbf{x}_i$, to another, $\mathbf{x}_j$, when generating motor
commands according to the policy $\pi$.  Additionally, we let the probability
distribution depend on a latent variable $\mathbf{e}_k$, which represents the
current ``agent-environment configuration'': it summarizes all external factors
that influence the outcome of the agent's actions. For example, imagine a robot
with a control policy to grasp a bottle: executing this policy will obviously
only have a chance of success if there is a bottle in reach of the robot in the
current situation. As another example, imagine two identical robots, standing
in a corridor in front of two identical looking doors, one leading to a
kitchen, the other leading to a dining room. The sequence of observations that
the two robots would make when opening and passing through the respective doors
would of course be very different (one would probably see a fridge, while the
other would probably see a dining table). This environmental influence, or
sensorimotor context, is summarized in an abstract way by the variable
$\mathbf{e}_k$. In the above example
(Figure~\ref{subfig:ex_sm_interaction_1b}), $\mathbf{e}_k$ could be said to
assume one of two values, corresponding to one of the plotted curves,
respectively.

One might be inclined to think of the variable $\mathbf{e}_k$ as the set of all
possible states of the entire universe. But this would of course be entirely
misleading: instead, it is helpful to conceive of it as the most compact way to
encode all qualitatively different situations the agent can face, given its
sensorimotor apparatus. From this perspective, we can argue that a sensorimotor
contingency corresponds to a property of the agent's interaction with the
world, when the value of the latent variable $\mathbf{e}_k$ is fixed. Taking
again the example of object perception, we could say that a given
$\mathbf{e}_k$ corresponds to an agent-environment configuration where a
certain object is in front of the agent (as compared to the object for example
being in a certain absolute position in ``world coordinates'').

Thus, since $\mathbf{e}_k$ is a latent variable, the agent cannot easily
estimate the above probability. What it does observe as it explores its
sensorimotor space for a longer period of time (i.e. for large $n$) is the
marginal probability distribution
\begin{multline}
    Pr( X_{t+1} = \mathbf{x}_j ~ | ~ X_t = \mathbf{x}_i, \pi ) \\ = \sum_{\mathbf{e}_k} Pr( X_{t+1} = \mathbf{x}_j ~ | ~ X_t = \mathbf{x}_i, \pi, E_t = \mathbf{e}_k ).
\label{eq:marginalized_distribution}
\end{multline}
The agent therefore has to demarginalize this observed state transition
distribution by estimating the latent environmental causes of sensorimotor
states.

\label{sec:temporal_adjacency_of_contingencies}

One way that the agent can estimate the current state $\mathbf{e}_k$, once it
has achieved a suitable demarginalization, is by tracking the likelihood of the
individual models (corresponding to the individual terms of the sum in
Equation~\ref{eq:marginalized_distribution}) over time, that is, using the
models to make predictions and comparing each prediction with actually observed
future states and updating model likelihoods correspondingly. Importantly, this
implies that the distinguishing property on which the agent relies to decide
whether two sensorimotor states are similar or not is \emph{temporal
adjacency}: given a high likelihood of an internal model, corresponding to a
hypothesis about a contingency, the agent assumes that the next observation
will also originate from the same contingency, unless the agent itself
disengages from it.  Furthermore, the agent can \emph{test} whether two
sensorimotor states are part of the same contingency by trying to reach the one
state from the other, by means of its own actions. If the agent can reliably
transition between two states whenever a certain internal model has a high
likelihood, it can safely assume that both states are part of the corresponding
contingency and can update the model accordingly.  On the contrary, if the
agent cannot reach some sensorimotor state, it can be inferred that this state
does not belong to the current contingency.

In the following, we will describe how we can utilize this property to
formulate an algorithm to demarginalize the transition probability distribution
and to discover latent structure, for an agent to learn predictive models of
sensorimotor contingencies.

\subsection{Contingencies as Clusters of Mutually Reachable States}

\label{subsec:t_as_adjacency}

Equation~\ref{eq:marginalized_distribution} describes the marginal probability
distribution of state transitions, which an agent observes when following a
policy $\pi$ for an extended amount of time. In the discrete case, this
distribution can be captured in a transition probability matrix $T_{\pi}$
with entries
\begin{equation}
    \left(T_{\pi}\right)_{i, j} = \Pr( X_{t+1} = \mathbf{x}_j ~ | ~ X_t = \mathbf{x}_i, \pi ),
    \label{eq:transitionprobabilitymatrix}
\end{equation}
where $\mathbf{x}_{i}, \mathbf{x}_{j} \in \mathcal{X}$.

As argued above, sensorimotor states that belong to the same contingency share
the property that the agent can transition between the states using its own
actions, or in other words, they are ``reachable'' from one another for the
agent. In contrast, states that are not reachable in this sense do not belong
to the same sensorimotor contingency.

The task to discover contingencies within the flux of sensorimotor states can
thus be reformulated as one of finding sets of sensorimotor states, for which
the agent observes a high probability of transitioning \emph{within} the set
while following its policy, but observing transitions \emph{out of} the set
only with a low probability. Note that it is not generally possible to
construct such sets by grouping together sensorimotor states based on how
``similar'' they are in sensorimotor space (measured for example by their
Cartesian distance). While a state $\mathbf{x}_i \in \mathcal{X}_p$ might have
a low distance to another state $\mathbf{x}_j \in \mathcal{X} \setminus
\mathcal{X}_p$, they can be dissimilar from a reachability point of view: they
do not co-occur in the same contingency (meaning for example that the agent
cannot reach $\mathbf{x}_j$ from $\mathbf{x}_i$ using its own actions). We will
see examples for such cases later in the descriptions of the simulation
experiments. A consequence of this property is that standard clustering
methods, such as K-means, will not find a valuable solution.

When interpreting the transition probability matrix $T_{\pi}$ as an adjacency
matrix for a weighted graph, where nodes represent sensorimotor states and
edges represent transition probabilities, then the task of discovering
sensorimotor contingencies consequently translates into one of finding densely
connected subgraphs within this graph, or said otherwise, finding a partition
that separates graph clusters. A suitable partition $\mathcal{P} = \{
\mathcal{X}_1, \mathcal{X}_2, \dots, \mathcal{X}_k \}$ of $\mathcal{X}$ (the
nodes of the graph) into $k$ subsets, which satisfies the property of having
high intra-subset transition probabilities and low transition probabilities in
between subsets can be found with the following minimization:
\begin{eqnarray}
\underset{\mathcal{P}}{\arg\min}
    \sum_{\mathcal{X}_p \in \mathcal{P}} \sum_{
        \substack{
            \mathbf{x}_i \in \mathcal{X}_p \\
            \mathbf{x}_j \in \mathcal{X} \setminus \mathcal{X}_p
        }
    } \Pr( X_{t+1} = \mathbf{x}_j ~ | ~ X_t = \mathbf{x}_i, \pi ) \\
=\underset{\mathcal{P}}{\arg\min}
    \sum_{\mathcal{X}_p \in \mathcal{P}} \sum_{
        \substack{
            \mathbf{x}_i \in \mathcal{X}_p \\
            \mathbf{x}_j \in \mathcal{X} \setminus \mathcal{X}_p
        }
    }     \left(T_{\pi}\right)_{i, j}.
    \label{eq:spectralclusteringcost}
\end{eqnarray}
Such a partition should constitute a solution to the problem of the agent of
discovering sets of states that correspond to different latent causes. Thus,
given a suitable partition $\mathcal{P}$, the probability distribution in
Equation~\ref{eq:state_transition_probability} can be estimated as
\begin{multline}
    \Pr( X_{t+1} = \mathbf{x}_j ~ | ~ X_t = \mathbf{x}_i, \pi, E_t = \mathbf{e}_k ) \\
    \sim \Pr( X_{t+1} = \mathbf{x}_j ~ | ~ X_t = \mathbf{x}_i, \pi, \mathbf{x}_i \in \mathcal{X}_k ),
\label{eq:model_estimation}
\end{multline}

\section{Computational Model}
\label{sec:computational_model}

\label{sec:spectral_clustering}

The minimization problem defined in Equation \ref{eq:spectralclusteringcost} is
an instance of the \emph{mincut} problem of cutting the transition probability
graph (which has the transition probability matrix $T_{\pi}$ as its adjacency
matrix) in such a way into $k$ components that the cut-crossing edges have
minimal weight. Each of the components would correspond to sets of sensorimotor
states with high ``inter-reachability'', whereas transitions in between sets
occur less frequently, thus matching the property of contingencies when viewed
from a graph point of view as described above.

The idea to use graph clustering methods to partition a state graph has already
been suggested in the reinforcement learning literature
\cite{csimcsek2005identifying, mannor2004dynamic}, but with a different
motivation: the aim of these works is to discover ``subgoals'', to speed up
learning convergence in reinforcement learning (see also
\cite{barto2003recent}). The idea is that ``bottlenecks'' (such as doorways in
a navigation task) are important subgoals when discovering a policy, and they
can be characterized as state transitions with low probability between two
clusters of densely connected states. Such transitions can be discovered by
formulating a similar minimization problem as the one proposed above. In
contrast, here we are interested in discovering densely connected clusters,
which we interpret as sensorimotor contingencies.

We thus want to solve the \emph{mincut} problem for the transition probability
graph defined by the matrix $T_{\pi}$ to find a partition into $k$ components.
This can be approximately achieved through spectral clustering
\cite{von2007tutorial}, using the method suggested by Ng and colleagues
\cite{ng2002spectral}, in the following way.

We first solve the eigenvalue problem for the transition probability matrix to
find the $k$ largest eigenvalues $\lambda_1, \lambda_2, \dots, \lambda_k$ and
associated eigenvectors $u_1, u_2, \dots, u_k$, and form the matrix
\begin{equation}
U = [u_1~u_2~\dots~u_k] \in \mathbb{R}^{m \times k}
\label{eq:eigenvectormatrix}
\end{equation}
where $m = |\mathcal{X}|$ is the number of discrete sensorimotor states that
the agent can observe. The matrix $U$ is then normalized such that each row has
unit length, resulting in the matrix $V$ with entries
\begin{equation}
V_{i, j} = \frac{U_{i, j}}{\sqrt{ \sum_l U_{i, l}^2 }}.
\label{eq:normalizedeigenvectormatrix}
\end{equation}

Treating each row in $V$ as a point in $\mathbb{R}^{k}$, we cluster them into
$k$ clusters using K-means. Since each row in $V$ also corresponds to a row in
the original transition probability matrix $T_{\pi}$, and thus to a state
$\mathbf{x} \in \mathcal{X}$, the result of the spectral clustering provides us
with the final partition of $\mathcal{X}$ into the $k$ subsets $\{
\mathcal{X}_1, \mathcal{X}_2, \dots, \mathcal{X}_k \}$. The agent can then use
this partition of states to train multiple internal models, one for each subset
$\mathcal{X}_i$ (see Equation~\ref{eq:model_estimation}).

\section{Simulation Experiments}
\label{sec:simulation_1}

\begin{figure}
\begin{center}
\includegraphics[width=\linewidth]{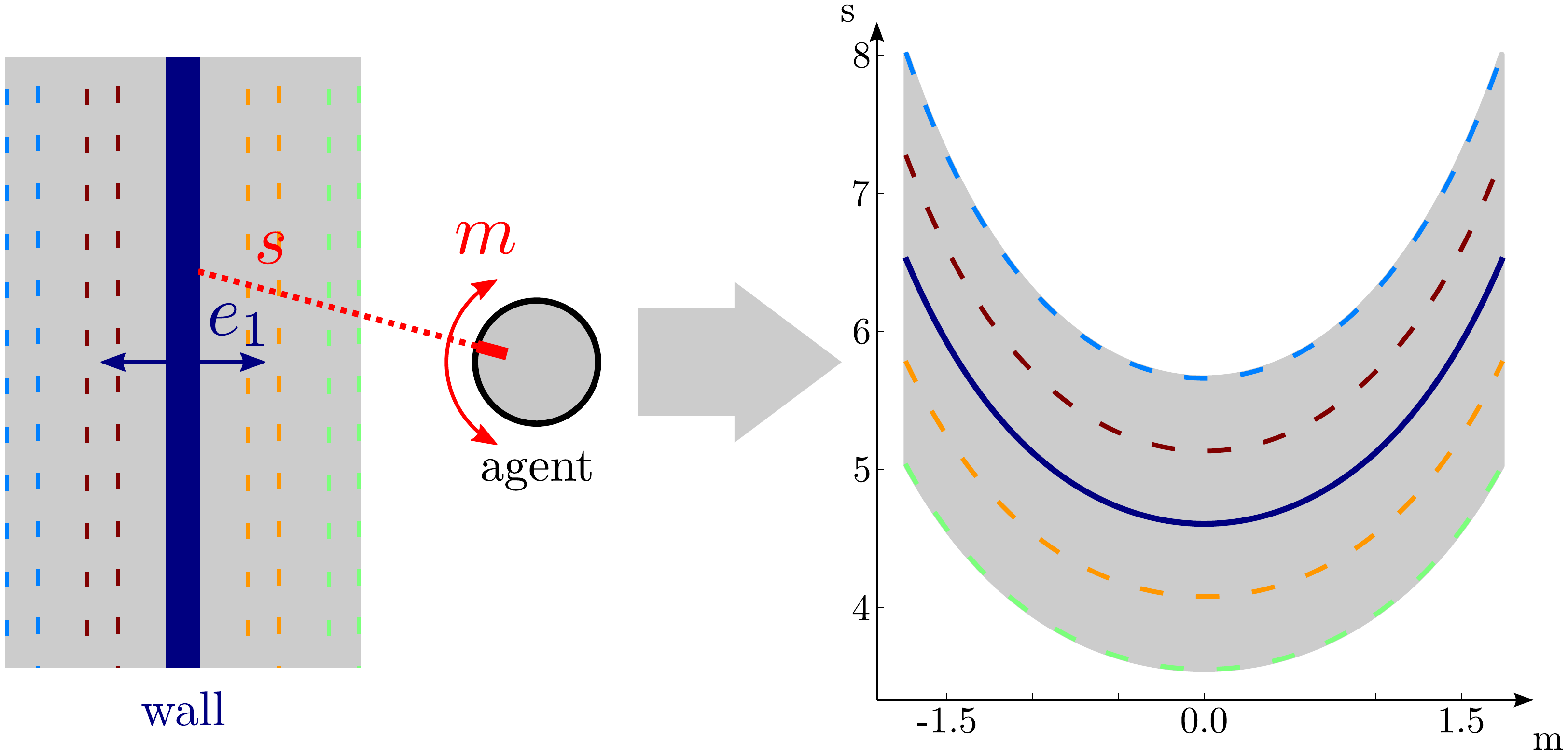}
\end{center}
\vspace{-1.5\baselineskip}
\setcounter{subfigure}{0}
\subfloat[simulated world: wall translates]{\label{subfig:robotscenario_world_1d}\phantom{\hspace{10em}}}
\hspace{4em}
\setcounter{subfigure}{2}
\subfloat[sensorimotor space]{\label{subfig:robotscenario_sms_1d}\hspace{10em}}
\begin{center}
\includegraphics[width=\linewidth]{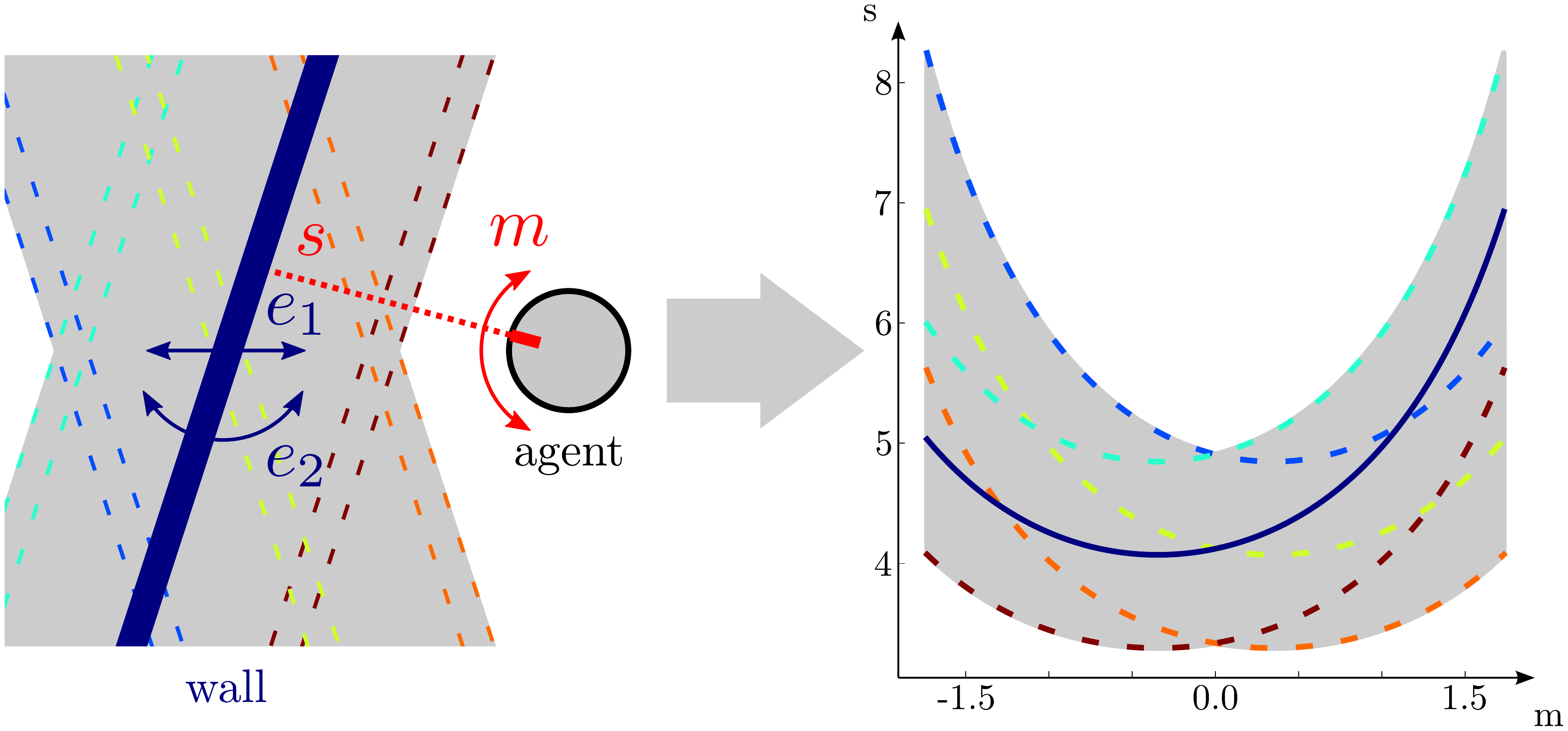}
\end{center}
\vspace{-1.5\baselineskip}
\setcounter{subfigure}{1}
\subfloat[simulated world: wall translates and rotates]{\label{subfig:robotscenario_world_2d}\phantom{\hspace{10em}}}
\hspace{4em}
\setcounter{subfigure}{3}
\subfloat[sensorimotor space]{\label{subfig:robotscenario_sms_2d}\hspace{10em}}
\caption{The simulation scenario that was used in the experiments. \protect\subref{subfig:robotscenario_world_1d} In the first variant, a wall is placed in front of the robot and moves towards and away from it. Several possible wall positions are indicated in different colors. \protect\subref{subfig:robotscenario_world_2d} In the second variant, the wall also rotates. In \protect\subref{subfig:robotscenario_sms_1d}--\protect\subref{subfig:robotscenario_sms_2d} the sensorimotor space is shown for the two variants, with observed manifolds plotted in color, corresponding to the shown wall configuration examples.}
\label{fig:robotscenario}
\label{fig:sensorimotorenvironment}
\label{fig:samplingwithouttime}
\end{figure}

\begin{figure*}
\begin{center}
\phantom{.}
\hfill
\subfloat[Wall translating, five discrete positions.]{  
\includegraphics[width=0.2\linewidth]{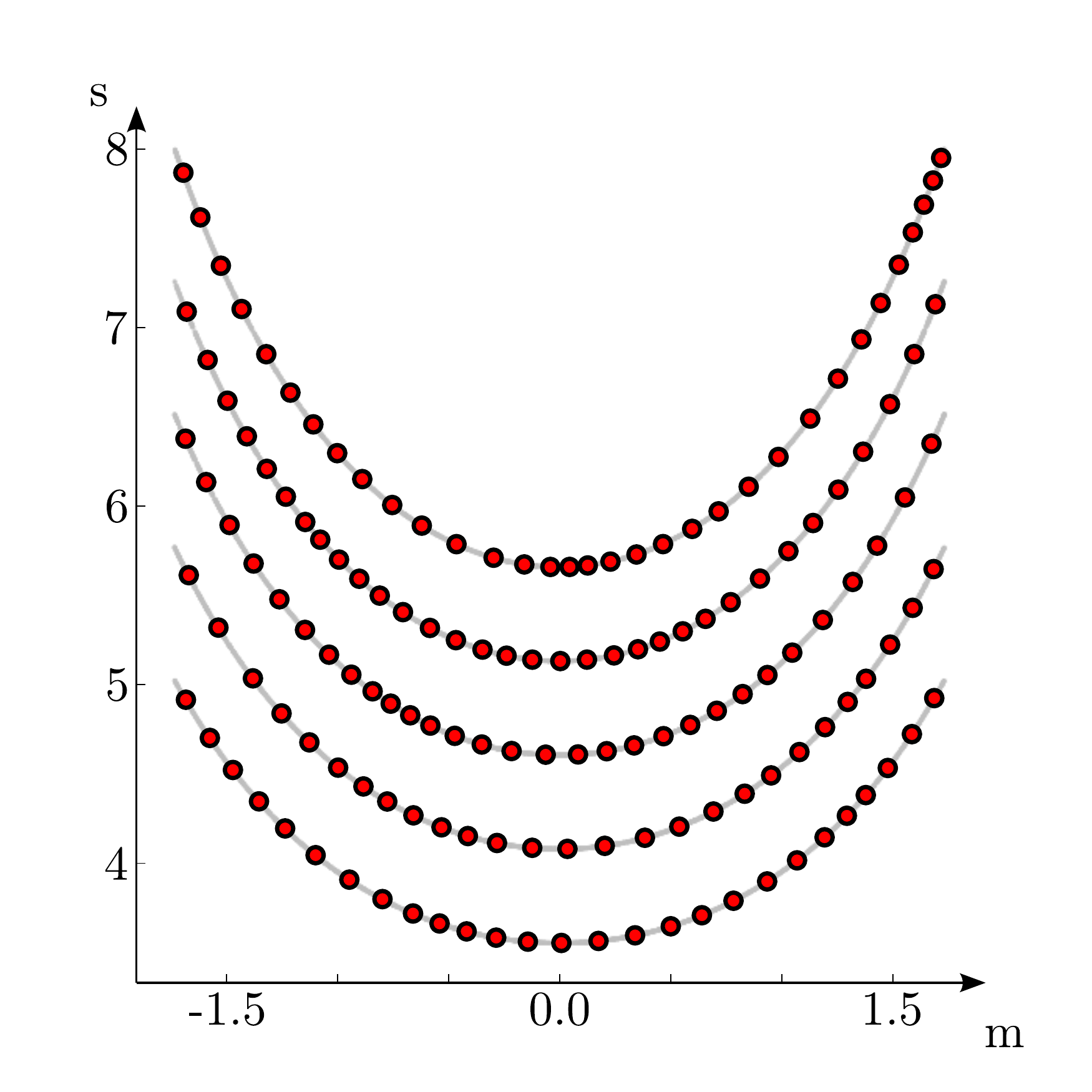}
\label{subfig:kmeans_5x1}
}
\hfill
\subfloat[Wall translating, continuous set of positions.]{  
\includegraphics[width=0.2\linewidth]{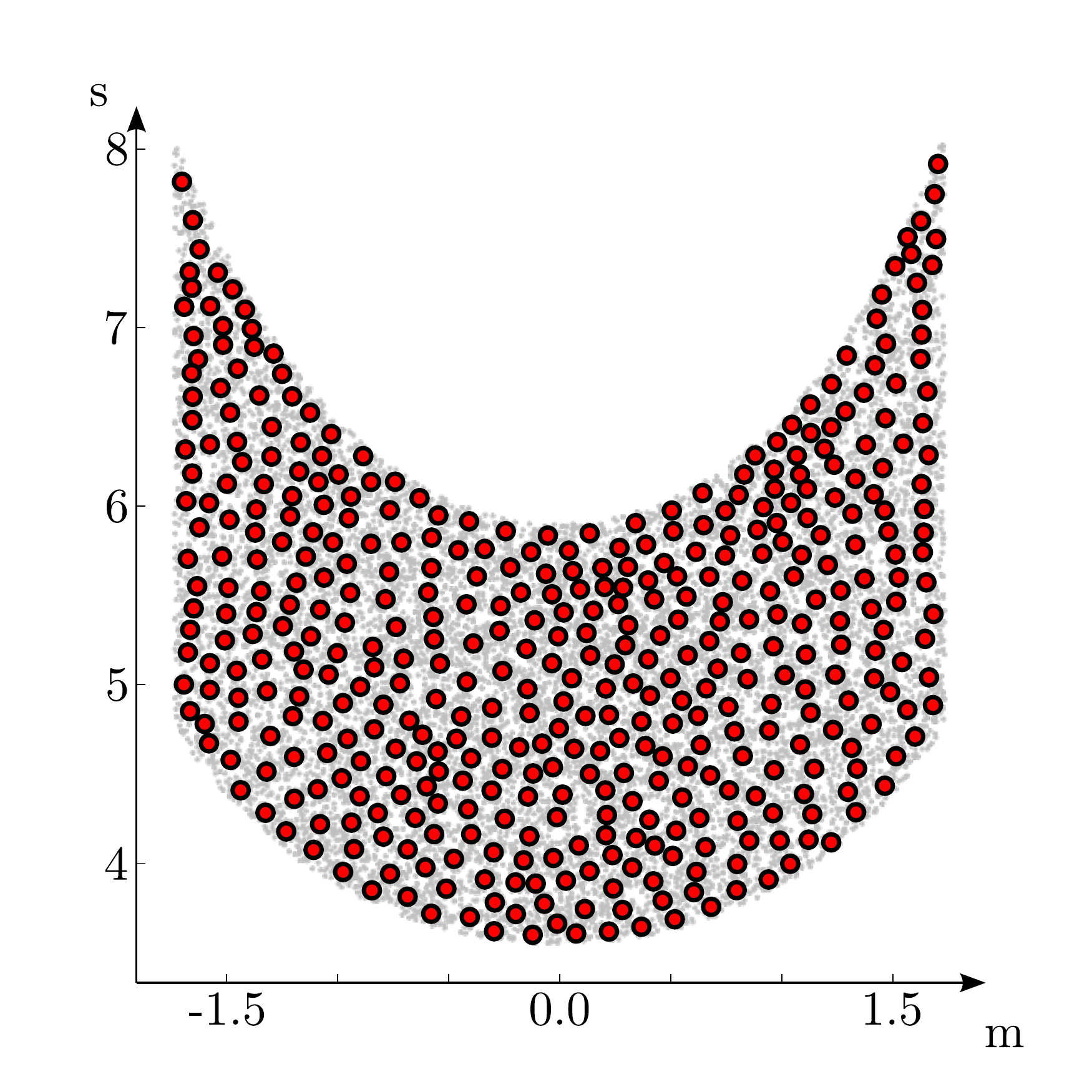}
\label{subfig:kmeans_1dc}
}
\hfill
\subfloat[Wall translating and rotating, six discrete positions.]{  
\includegraphics[width=0.2\linewidth]{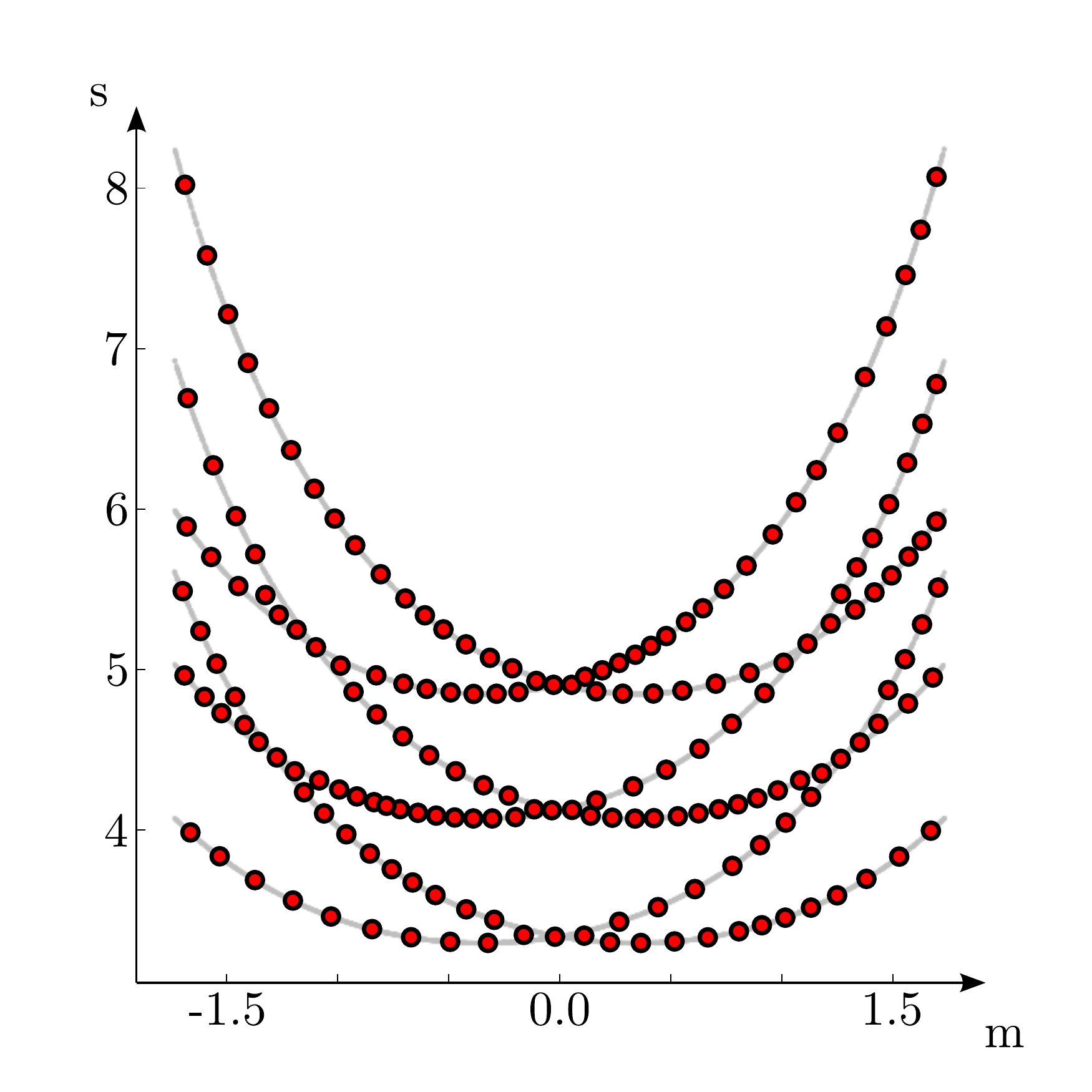}
\label{subfig:kmeans_3x2}
}
\hfill
\phantom{.}
\end{center}
\caption{The outcome of the agent's initial exploration of the sensorimotor space, in three variants of the simulation experiment. Sensorimotor observations are plotted as gray dots, cluster centroids are shown as red circles.} 
\label{fig:kmeans}
\end{figure*}

To evaluate the method proposed above, a simplistic simulation scenario was
designed with the purpose of highlighting several important aspects of the
learning problem that a naive agent is facing when learning to interact with
its environment. Consider the situation of an agent (a virtual robot) with a
turnable base and a single laser range sensor, measuring the distance to the
nearest obstacle in a straight line of sight (see
Figures~\ref{fig:robotscenario}\subref*{subfig:robotscenario_world_1d}--\subref*{subfig:robotscenario_world_2d}).
In front of the agent, in the otherwise empty environment, a movable wall is
placed. As the agent explores its situation by generating motor commands, i.e.
rotating around its base, it makes a sequence of observations.  As long as the
wall remains static, these observations all lie on a one-dimensional manifold
in the agent's two-dimensional sensorimotor space.  However, in this scenario,
the wall is randomly moved into new positions. As the wall changes positions,
also the manifold in sensorimotor space changes, on which the observations lie
that the agent makes. This can be seen in
Figures~\ref{fig:robotscenario}\subref*{subfig:robotscenario_sms_1d}--\subref*{subfig:robotscenario_sms_2d},
which show the agent's two-dimensional sensorimotor space (with the motor
degree of freedom along the horizontal axis, and the sensory input along the
vertical axis): for each distinct state of the world (i.e. position and
orientation of the wall), the agent's observations lie on a different curve.

In this simulation scenario, the latent environmental state $\mathbf{e}_k$ (cf.
Equation~\ref{eq:state_transition_probability}) consists of the degrees of
freedom of the wall: its position and orientation. For each distinct state, the
agent's exploration results in observations lying on a corresponding distinct
manifold. When considering this process in terms of the marginal probability
distribution described by Equation~\ref{eq:marginalized_distribution}, each
individual state of the environments corresponds to one of the summands, and
thus a distinct distribution.

The agent however only has access to its own sensor and motor states, whereas
the state of the wall is unknown to it. It also has no information about when
the environment changes states. But the sensory feedback that it receives when
exploring its environment is directly influenced by the state of the
environment. From an external perspective, the observations made by the agent
can be characterized as sequences of curve segments, a new segment beginning
each time when the environment changes states. However, the agent a priori does
not know about the existence of this structure in the data. It experiences
observations distributed in sensorimotor space, without knowing that they
belong to distinct subspaces, or manifolds. From the agent's point of view, the
samples are generated from the marginal probability distribution of
sensorimotor state transitions, see
Equation~\ref{eq:marginalized_distribution}.  When trying to directly estimate
a model of the sample distribution from all the samples taken together, the
agent would thus face the difficulty of having a large sensory variance
associated with each transition, which in turn would result in a low accuracy
for prediction.

However, as suggested above, the agent can solve this problem by clustering
observations together in terms of how well it can \emph{reach} them from one
another, as opposed to how similar they \emph{appear} (as for example measured
by their Cartesian distance in sensorimotor space), thereby demarginalizing the
probability distribution, allowing the agent to estimate multiple separate
internal models, to account for different latent states.

In a first step, the continuous sensorimotor space of the agent is discretized
in an initial sensorimotor exploration phase, to construct the set of
sensorimotor states $\mathcal{X}$: The agent generates observations by
exploring its motor space and observing sensory responses, across a timespan
long enough for the environment to visit all of its states.  In the
experiments, a Gaussian random walk was used as exploration policy $\pi$ for
the agent's motor outputs while the environment's state changed at each time
step with a probability of $0.1$ (also performing a Gaussian random walk in its
own state space $\mathcal{E} = \left\{\mathbf{e}_i\right\}$), thus allowing the
agent to explore each constant state of the environment on average for 10 time
steps. K-means is then applied on the set of observations to obtain a
discretization of the sensorimotor space into $r$ states, resulting in the set
of sensorimotor states $\mathcal{X} = \{\mathbf{x}_i\}$, $i = 1, \dots, r$,
$\mathbf{x}_i = (m_i, s_i)^T$. The outcome is plotted in
Figures~\ref{fig:kmeans}\subref*{subfig:kmeans_5x1}--\subref*{subfig:kmeans_3x2}
for different versions of the scenario.  Subsequently, sensorimotor
observations are classified as one of these states using nearest-neighbor
classification.

\begin{figure*}
\begin{center}
\includegraphics[width=\linewidth]{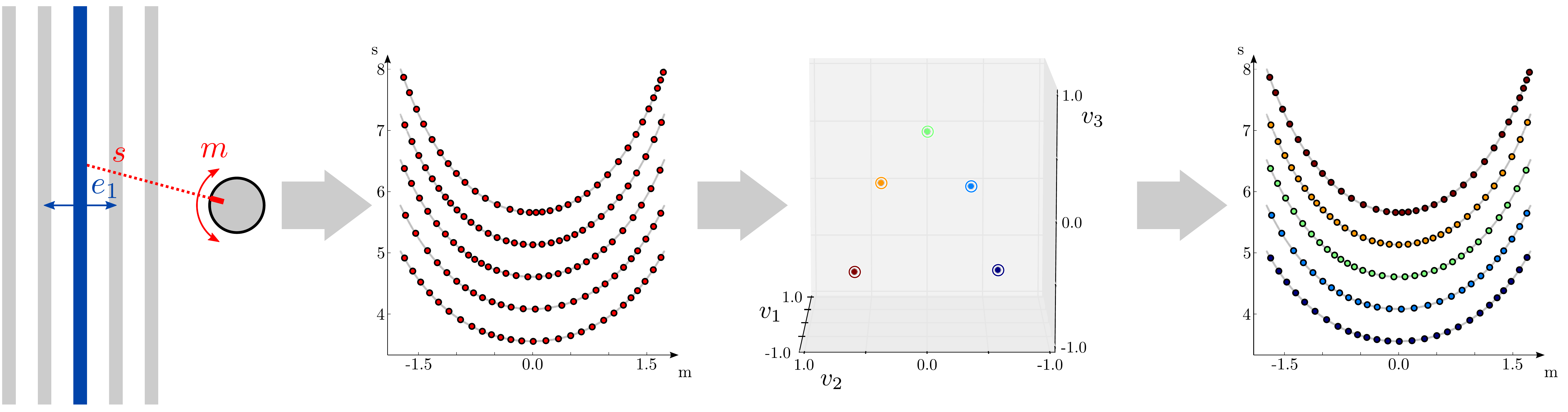}
\end{center}
\vspace{-2\baselineskip}
\subfloat[simulated world]{\label{subfig:simulation_1a}\phantom{MMMMMMMM}}
\hspace{6em}
\subfloat[sensorimotor space]{\label{subfig:simulation_1b}\phantom{MMMMMMMMM}}
\hspace{5.5em}
\subfloat[spectral space]{\label{subfig:simulation_1c}\phantom{MMMMMMM}}
\hspace{7.5em}
\subfloat[sensorimotor contexts]{\label{subfig:simulation_1d}\phantom{MMMMMMMMMM}}

\caption{Overview of Simulation 1. \protect\subref{subfig:simulation_1a} The
wall moves into five distinct positions, shown in gray. The five distinct
configurations can also clearly be seen in the agent's sensorimotor space
\protect\subref{subfig:simulation_1b}, as well as in its spectral projection
\protect\subref{subfig:simulation_1c}. In the latter, the outcome of the
spectral clustering is indicated by the different colors, showing that
\protect\subref{subfig:simulation_1d} the five sensorimotor contexts have
correctly been discovered.}

\label{fig:simulation_1}
\end{figure*}

\begin{figure}
\begin{center}
\includegraphics[width=\linewidth]{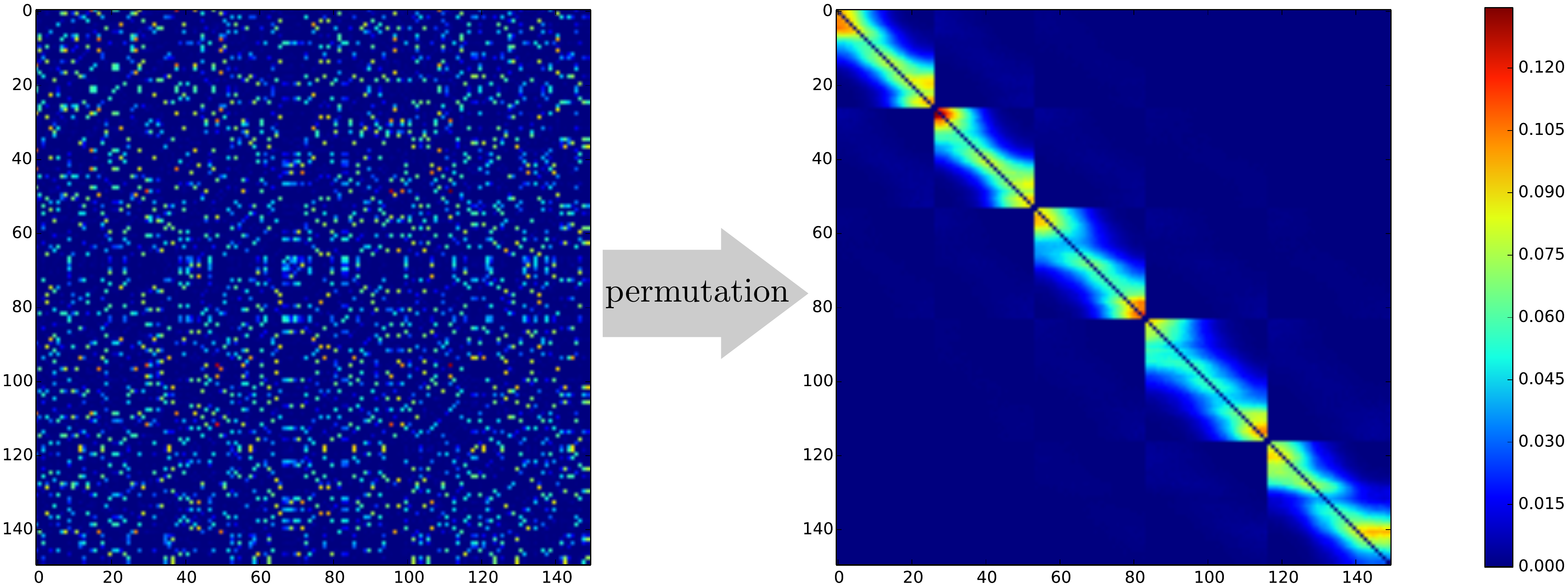}
\end{center}
\caption{Via a permutation of the transition probability matrix $T_{\pi}$, some
of the underlying structure of the sensorimotor interaction of the agent with
its environment becomes apparent.}
\label{fig:tpm_reordering}
\end{figure}

The agent then continues to explore its sensorimotor space for a fixed amount
of time and thus observes a sequence of sensorimotor states. Above it was
argued that sensorimotor contingencies are characterized by sets of states that
are ``inter-reachable'' for the agent, meaning that it can transform any state
within the set into any other state from the same set, with its own actions.
The agent thus only requires to retain information about state transitions:
knowledge about which state transitions are possible is sufficient to derive
which states can be reached from one another. We therefore remove from the
sequence of observed sensorimotor states all duplicate entries, such that we
obtain a sequence of states (cf. Equation~\ref{eq:stochastic_process}) in which
each pair of subsequent elements correspond to an observed state transition:
\begin{equation}
X^{\pi} = \left\{ X_t = \mathbf{x}_i~:~t = 1, \dots, n \right\},~\textrm{s.t.}~\forall~t:~X_t \neq X_{t+1}.
\end{equation}
Based on these observed state transitions, the transition probability matrix
$T_{\pi}$ is then constructed (cf.
Equation~\ref{eq:transitionprobabilitymatrix}), where each row is normalized to
sum up to $1$ to form a probability mass function. Spectral clustering is then
applied as per the method presented in Section~\ref{sec:spectral_clustering}
to obtain a partition of the set of sensorimotor states.

In the following, several simulation runs will be presented, to visualize the
result of the learning method, and to point out properties of the learning
problem that a naive agent is facing when trying to learn sensorimotor
contingencies as described above.

\subsection{Simulation 1: Spectral Transform of Sensorimotor States}

\begin{figure*}
\begin{center}
\includegraphics[width=\linewidth]{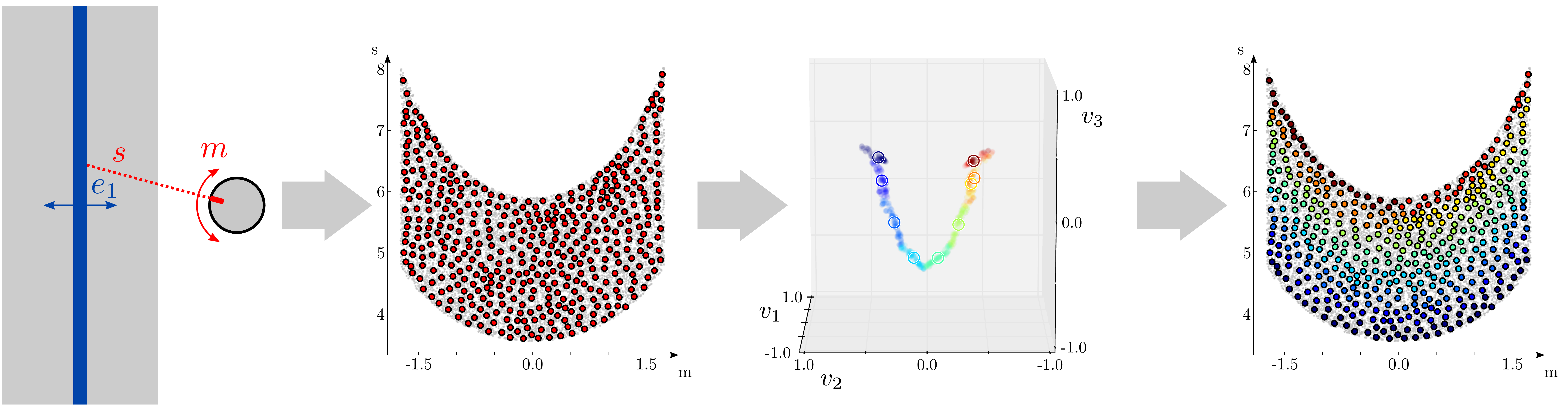}
\end{center}
\vspace{-2\baselineskip}
\subfloat[simulated world]{\label{subfig:simulation_2a}\phantom{MMMMMMMM}}
\hspace{6em}
\subfloat[sensorimotor space]{\label{subfig:simulation_2b}\phantom{MMMMMMMMM}}
\hspace{5.5em}
\subfloat[spectral space]{\label{subfig:simulation_2c}\phantom{MMMMMMM}}
\hspace{7.5em}
\subfloat[sensorimotor contexts]{\label{subfig:simulation_2d}\phantom{MMMMMMMMMM}}

\caption{Overview of Simulation 2. \protect\subref{subfig:simulation_2a} In
this variant, the wall moves freely within an interval of possible positions
(shown as a gray area). \protect\subref{subfig:simulation_2b} Distinct
manifolds are no longer visible in sensorimotor space, but
\protect\subref{subfig:simulation_2c} its projection into spectral space
results in a distribution of the sensorimotor states along a one-dimensional
manifold, capturing the one degree-of-freedom of the environment.
\protect\subref{subfig:simulation_2d} The spectral clustering successfully
captures the structure of the interaction: discovered sensorimotor contexts
match the curvature of the continuum of manifolds.}

\label{fig:simulation_2}
\end{figure*}

Consider first the case of the simulation scenario where the wall only moves
towards and away from the agent into five distinct states, while its
orientation is held constant (see Figure~\ref{fig:simulation_1} for an overview
of the simulation experiment). Figure~\ref{fig:tpm_reordering} shows the
transition probability matrix $T_{\pi}$ that was constructed in this case. We
can uncover some of the structure of the interaction by permuting the rows and
columns: Since the generation of sensorimotor states through K-means does not
carry any information about the topology of the sensorimotor space, the
ordering of rows and columns is initially random. However, for the purpose of
visualization, we can permute them to match the true topology of the data. This
way it becomes easily visible that each of the states of the wall manifests
itself for the agent as a ``cluster'' of states between which the agent can
transition with high probability, whereas transitions between the clusters have
a low probability.  This can be seen in the figure as blocks of entries with
high values along the diagonal of the matrix, one for each of the states of the
environment. Note that the fact that some of the sensorimotor states have a
higher probability to be transitioned to, in particular ones lying close to the
left and right limits of the motor space, is due to the way that cluster
centroids were placed in sensorimotor space by the K-means algorithm: because
of the curvature of the manifolds, the centroids are more ``packed'' towards
the center of the motor space and more spread apart towards the ends, rendering
it overall more likely for the agent to observe the centroids lying close to
the ends of the motor space.

When interpreting $T_{\pi}$ as an adjacency matrix for a graph, as suggested
above in Section~\ref{subsec:t_as_adjacency}, each of these blocks corresponds
to a densely connected subgraph within this graph. Note that the reordering has
no effect whatsoever on the result of the learning method, and is therefore
unnecessary.  It was only done here once for the purpose of visualization, and
normally is not part of the learning method.

In the spectral clustering step, the matrix $V \in \mathbb{R}^{m \times k}$ is
constructed based on an eigenvalue decomposition of the matrix $T_{\pi}$, as
described in Section~\ref{sec:spectral_clustering}. Each of the $k$ rows in $V$
corresponds to exactly one sensorimotor state $\mathbf{x}_i$, and is treated as
a coordinate in an $m$-dimensional ``spectral space''. The $m$ dimensions of
the spectral space in turn correspond to the eigenvectors associated to the $m$
strongest eigenvalues of $T_{\pi}$. To visualize this mapping of sensorimotor
space into the spectral space, Figure~\ref{subfig:simulation_1c} shows a plot
of a projection of the $k$ points into the three-dimensional subspace of the
three strongest eigenvalues: it can be seen that all points lie in one of five
dense clusters. The actual clustering is performed using K-means in the
$m$-dimensional spectral space, where $m$ is also the number of clusters to be
found.

Selecting the correct number of clusters $m$ of course plays an important role
in the outcome, as in any clustering method. However, for the purpose of this
discussion, the problem of how to estimate $m$ is not the focus and instead a
good value is manually selected. However it should be noted that methods exist
to automatically estimate the number of clusters \cite{jain2010data}. In this
version of the scenario, it is set to $m=5$, to match the number of discrete
positions that the wall is entering. Figure~\ref{subfig:simulation_1d} shows
the result of the clustering of the sensorimotor states: it can be seen that
indeed each of the five dense clusters in spectral space groups together all
sensorimotor states that belong to one of the position of the wall.

\subsection{Simulation 2: The Topology of the Internal State Representation}

We can learn more about the topology of the distribution of sensorimotor states
in spectral space when considering another variant of the scenario: in this
variant, the wall changes positions in form of a continuous Gaussian random
walk. The latent variable $\mathbf{e}$ therefore assumes values from an
interval, instead of only five discrete values as before.
Figure~\ref{fig:simulation_2} shows an overview of this version of the
simulation experiment. Here it can be seen, that the states are no longer
clearly separated in spectral space, but instead have a curve-like
distribution. Furthermore, as indicated by the color code, this curve-like
distribution can be seen as capturing the one degree-of-freedom of the
environment: Sensorimotor states that lie at the bottom of sensorimotor space
(i.e. states that are observed when the wall is in the position that is closest
to the agent) have been mapped to the left end of the curve. As we move
rightwards along the curve, the positions of states move upwards in
sensorimotor space, until finally, at the other end of the curve, all states
are gathered that correspond to the topmost manifold in sensorimotor space
(i.e. states that are observed when the wall is in the position that is
farthest away from the agent). Note that what is shown in the figure is a
three-dimensional projection of a higher-dimensional distribution, in this case
$m=10$. Because of the normalization that is done in
Equation~\ref{eq:normalizedeigenvectormatrix}, the points actually lie on the
surface of an $m$-dimensional hypersphere.

The result of applying spectral clustering in this case is shown in
Figure~\ref{subfig:simulation_2d}. It can be seen that overall, sensorimotor
states that are located close to similar manifolds (i.e. wall positions) have
been clustered together. Note however, that the top-most clusters are ``broken
apart'' in the center: this is a result of choosing a larger value for the
parameter $m$ in the K-means step. These clusters that represent curve segments
could be merged together in a subsequent step, to improve the quality of the
result. How and \emph{why} the agent could do this will be discussed in the
conclusion of this article.

To understand why the sensorimotor points are distributed in a curve-like way
in spectral space, that is, along a one-dimensional manifold on the surface of
the $m$-dimensional hypersphere, it is again helpful to think of the matrix
$T_{\pi}$ as an adjacency matrix of a graph. The construction of the spectral
space can then be seen as a means to map out the graph in $m$-dimensional
space, such that nodes (i.e. sensorimotor states) that are connected via edges
with high weights are placed close together. Transitions between sensorimotor
states with higher probability can thus be seen as ``springs'' that pull these
states together in spectral space. In the case of the first simulation, the
result is quite clear: only transitions within one of the five clusters had
relatively high probabilities, and thus the states of each of the clusters were
``pulled together'', effectively collapsing them into five almost point-like
distributions. In the case of the second simulation however, the situation is a
bit different: Each sensorimotor state can now be observed not only in a
single, but many wall positions. Importantly, this means that for a given
sensorimotor state, the agent will observe transitions to different other
states, depending on the exact position of the wall (for example when moving
left in motor space, for one wall position it might transition ``left and
down'' to another state, whereas it might transition ''left and up'' when the
wall is slightly farther away). As a result, similar sets of sensorimotor
states are observed by the agent when the wall is in similar configurations,
and thus, they are arranged closely together in spectral space. On the other
hand, sensorimotor states that only occur in very different wall configurations
are not bound together by high transition probabilities, and thus are mapped
further apart in spectral space. This behavior results in the generation of
the point distribution that can be seen in Figure~\ref{subfig:simulation_2c}.

\begin{figure*}
\begin{center}
\includegraphics[width=\linewidth]{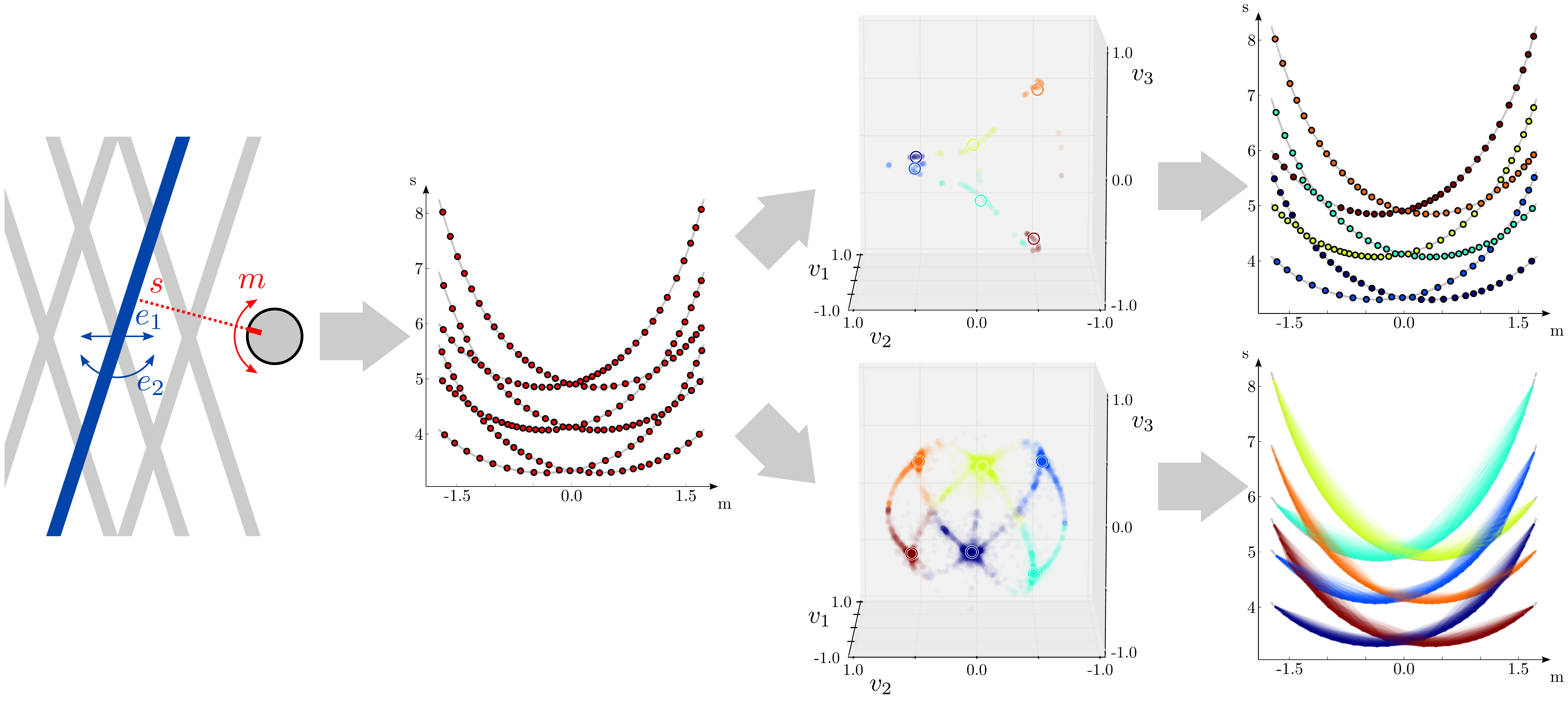}
\end{center}
\vspace{-2\baselineskip}
\subfloat[simulated world]{\label{subfig:simulation_3a}\phantom{MMMMMMMM}}
\hspace{7em}
\subfloat[sensorimotor space]{\label{subfig:simulation_3b}\phantom{MMMMMMMMM}}
\hspace{6.5em}
\subfloat[spectral space]{\label{subfig:simulation_3c}\phantom{MMMMMMM}}
\hspace{6em}
\subfloat[sensorimotor contexts]{\label{subfig:simulation_3d}\phantom{MMMMMMMMMM}}

\caption{Overview of Simulation 3. \protect\subref{subfig:simulation_3a} In
this variant, the wall not only translates but also rotates, resulting in
intersecting manifolds in sensorimotor space
\protect\subref{subfig:simulation_3b}. The upper and lower paths of
\protect\subref{subfig:simulation_3c}--\protect\subref{subfig:simulation_3d}
show the result of the spectral clustering for the cases of clustering
observations and transitions, respectively (see text for details).}

\label{fig:simulation_3}
\end{figure*}

As a first qualitative result, it can thus be said that through the presented
learning method, a naive agent constructs an internal representational space
(the spectral space), in which observations happen to be distributed along a
one-dimensional manifold. This one-dimensional manifold happens to correspond
nicely to the actual one-dimensional latent state space $\mathbf{e}$ of the
environment, even though at no point was any information about the latent
variable put into the learning system. Furthermore, the internal
representational space of the agent \emph{stabilizes} the agent's perception of
the world: instead of trying to directly build up a predictive model of its
sensorimotor interaction, which is difficult because of large variances in
sensory observations as discussed further above, the agent could first
construct the internal space and subsequently select subsets of observations,
based on which it could train individual predictive models. In a way this
allows the agent to virtually ``freeze'' the environment, to improve its
learning.

\subsection{Simulation 3: Ambiguity in Sensorimotor States}

So far, only simulations have been described in which the wall changed its
position, but not its orientation. What this means is that when the wall moves,
the manifold of observations in sensorimotor space translated parallel to the
axis corresponding to the sensory input dimension (the vertical axis in the
figures above). This in turn means that every point in the agents sensorimotor
space unambiguously belonged to exactly one manifold. The agent thus could
identify the latent state of the environment from a single observation, and
would be able to predict what other sensorimotor states to expect: those lying
on the same manifold.

In more technical terms, it can be said that the environment in the above
examples was Markovian: to make an accurate prediction about the next
sensorimotor state, the agent in principle only has to know the current
sensorimotor state. The assumption about the state distribution being Markovian
is often made in the reinforcement learning literature, but cannot be assumed
to be true in general. In fact, often a state space is specifically designed in
such a way that it is compact and casts the problem into a setting in which the
Markovian assumption holds.  But of course, this is not possible when the goal
is to develop \emph{generic} learning mechanisms, where the task knowledge of
the human designer cannot be incorporated into the learning system.

To see what situation a naive agent is facing when confronted with an
environment in which the Markovian assumption does not hold, consider the
variant of the simulation scenario where the wall not only moves, but also
rotates (see Figure~\ref{fig:simulation_3} for an overview of this third
simulation experiment). The key difference to the previous experiments, is that
the manifolds in sensorimotor space are now crossing (see
Figure~\ref{subfig:simulation_3b}). At each intersection point, the agent can
now no longer predict what other observations it will make, based solely on the
current observation. The sensorimotor states at the intersection points are
\emph{ambiguous}, belonging to multiple distinct contexts. This of course has
an important impact on the transition probabilities that the agent observes:
while transitions along non-intersecting curve segments have locally Markovian
distributions as in the previous simulations, transitions from intersection
points can end in sensorimotor states lying on any of the intersecting
manifolds.

When interpreting the mapping from sensorimotor space to spectral space again
in terms of a graph with ``spring-like'' edges, it becomes clear why the above
property of sensorimotor states lying on intersection points being connected to
other states from multiple manifolds has a strong impact on the spectral
mapping. This can be seen in the top plot of Figure~\ref{subfig:simulation_3c},
which shows again a three-dimensional projection of the (in this case
six-dimensional) spectral space. It can be seen that states belonging to
different contexts are no longer clearly grouped together in spectral space,
nor do they seem to be organized in a low-dimensional manifold.  Instead,
states from different contexts are ``pulled together'' by the ambiguous states,
resulting in a distribution that -- at least in the three-dimensional
projection -- appears to be ``folded'' in a non-trivial way.  The outcome of
the K-means clustering in spectral space, shown in the upper plot in
Figure~\ref{subfig:simulation_3d}, still correctly separates sensorimotor
states into different contexts. However, two things have to be noted: Firstly,
the clustering does not always produce the result that is shown here, but
instead sometimes incorrectly combines curve segments from intersecting
manifolds. This, as well, offers the interpretation that the distribution of
sensorimotor states is rather complex, not only in the three-dimensional
projection but also in the actual six-dimensional spectral space. Secondly, the
hard clustering that is produced by K-means of course results in each
sensorimotor state being associated to exactly one cluster, or context.
However, the sensorimotor states that lie at the intersection point between
manifolds should actually be considered as belonging to both of the
intersecting manifolds.

A first step towards resolving these issues is to change the representation of
the agent to the space of \emph{transitions} between observations. By doing
this, the method described above no longer clusters together observations, but
frequently co-occurring transitions between observations. A similar strategy
has also been successfully applied for example by Mnih et al. to disambiguate
observations \cite{mnih2013playing}. Minh et al. used sequences of video frames
as input to their reinforcement learning system instead of single frames, which
might be ambiguous.

The overall method remains unchanged, with the only adaptation being the notion
of state: not points in sensorimotor space, but transitions between points are
treated as states. To make the distinction explicit, a superscript arrow will
be used in the notation of this new definition of state:
$\mathcal{X}^{\rightarrow} = \{ \mathbf{x}^{\rightarrow}_{d_{i,j}}
\}$, where $d_{i,j}$ is an index, and the state
$\mathbf{x}^{\rightarrow}_{d_{i,j}} = (\mathbf{x}_i, \mathbf{x}_j)$ corresponds
to a transition from one K-means cluster, $\mathbf{x}_i$, to another,
$\mathbf{x}_j$. Consequently, the state transition probability (cf.
Equation~\ref{eq:state_transition_probability}) becomes
\begin{equation}
    \Pr( X_{t+1} = \mathbf{x}^{\rightarrow}_{d_{j, k}} ~ | ~ X_t = \mathbf{x}^{\rightarrow}_{d_{i, j}}, \pi, E_t = \mathbf{e}_l ),
    \label{eq:state_transition_probability_2step}
\end{equation}
and entries in the transition probability matrix $T^{\rightarrow}_{\pi}$ become
\begin{equation}
    \left(T^{\rightarrow}_{\pi}\right)_{d_{i,j}, d_{j,k}} = \Pr( X_{t+1} = \mathbf{x}^{\rightarrow}_{d_{j,k}} ~ | ~ X_t = \mathbf{x}^{\rightarrow}_{d_{i,j}}, \pi ).
    \label{eq:transitionprobabilitymatrix_2step}
\end{equation}

The lower parts of Figures~\ref{fig:simulation_3}\subref*{subfig:simulation_3c}--\subref*{subfig:simulation_3d} show the outcome of applying the algorithm when using this new definition of state. Each
point in spectral space in this case represents one transition between clusters in sensorimotor
space. Transitions within one manifold have again been nicely grouped together,
forming six clearly visible clusters of states that can easily be separated
using K-means.

\section{Conclusion}

Despite the impressive progress that has been made in machine learning research
recently, we still do not know how to efficiently implement the learning of
\emph{generic} knowledge in artificial agents, such as robots. This article
offered as a perspective on this topic to take inspiration from two recent
developments in the cognitive science literature: predictive processing
approaches to cognition, and the sensorimotor contingencies theory of
perception. More specifically, this paper is in line with recent works that
suggest a combination of PP and the SMCT \cite{seth2014predictive,
laflaquiere2015grounding}.

According to PP, the cognitive system of adaptive agents has to maintain a
hierarchical model of the latent causes of sensory observations. To answer the
question of how a naive agent could autonomously learn about these latent
causes, this article borrowed the concept of sensorimotor contingency from the
SMCT of perception: an agent should try to engage in predictable sensorimotor
interactions with its environment, by the means of which it perceives
sensorimotor contexts (or contingencies). This idea was formally cast into a
computational model as an instance of the \emph{mincut} problem on a graph of
sensorimotor state transitions, and spectral clustering was proposed as a
method to solve this problem.

In the proposed model, the agent discovers sensorimotor contexts as sets of
sensorimotor states that, for the agent, are reachable from one another: if it
observes one state and, after acting, observes another state with high
probability, then the agent can assume that these two states belong to the same
context. It was demonstrated in a series of simulation experiments that with the
proposed method, a naive agent can successfully discover sensorimotor contexts.

In the presented implementation of the method, the agent's knowledge about the
discovered sensorimotor contexts was represented in a very simple way: the set
of sensorimotor states was partitioned, resulting in a description of
sensorimotor contexts in the form of subsets of this set, in conjunction with a
transition probability matrix. As a next step, it would be possible to train
more accurate predictive models for each of the discovered sensorimotor
contexts, to improve the agent's capability to predict the flow of sensorimotor
observations. But already in this very simple implementation, the agent can be
said to reduce its overall prediction error: Were it to make predictions about
sensorimotor state transitions directly (i.e. without first discovering the
different sensorimotor contexts), it would have to base its predictions on the
estimate of the marginal probability distribution
(Equation~\ref{eq:marginalized_distribution}). In contrast, as discussed above,
the agent can demarginalize the distribution via the discovery of sensorimotor
contexts, effectively allowing it to estimate the latent variable $\mathbf{e}$
and to make its predictions under the posterior probability described by
Equation~\ref{eq:model_estimation}. The posterior probability distribution
necessarily has a lower entropy than the marginal probability distribution,
which is equivalent to saying that the agent improves its overall ability to
predict \cite{friston2007free}.

Two variants for defining the notion of ``sensorimotor state'' were used in the
simulation experiments: either prototype points in sensorimotor space (placed
using K-means), or transitions between such prototypes. As long as the space of
sensorimotor observations is unambiguous, the former definition is suitable for
the agent to recognize sensorimotor contexts. However, when sensorimotor
observations can belong to more than just a single context, this becomes
problematic, as was demonstrated in the simulation experiments. In that case,
ambiguous observations were arbitrarily assigned to one context. By implication
this means that the agent would always perceive these observations as belonging
to a certain context, even when they occur within another. It was demonstrated
that this problem could be solved by instead clustering transitions between
prototypes. In that case, the agent would no longer recognize a sensorimotor
context directly via the set of observations that occur within that context,
but via transitions between observations. Said otherwise, the agent has to act
in order to perceive a context. It should be noted however that it would still
not be strictly necessary for the agent to act to perceive anything. Instead,
when an observation unambiguously belongs to only one context (i.e. all
transitions from and to this observation belong to the same context), then
there is no reason to assume that the agent would first have to move to
perceive the context or contingency.

In all of the presented simulation experiments, the number of contexts to be
discovered was manually fixed by specifying the number of clusters that should
be found by the spectral clustering algorithm. Of course this should be done in
an autonomous way instead, which will be addressed in future work. One
possibility would be to use a method to estimate a suitable number of clusters
\cite{jain2010data}, as suggested above. However, it may be more natural to
approach this difficulty in the scope of the related question of how a system
should construct a representational hierarchy, in the following way. Initially,
a large number of clusters could be formed, resulting in a too finely grained
partition of sensorimotor states. But in a subsequent step, the next level of
the representational hierarchy could be constructed by reapplying the same
algorithm, but treating clusters as ``meta-states'': While the agent would
continue to interact with its environment, it would recognize observations as
belonging to a number of the initially discovered contexts, or meta-states,
namely those that together constitute the actual manifold in sensorimotor
space, on which the observations are distributed. It would therefore observe
\emph{transitions between these contexts} to occur with a high probability,
comparable to the high probability of transitions between sensorimotor states
that belong to the same manifold. The same process of discovering densely
connected subgraphs (by constructing a transition probability matrix and
applying the spectral clustering algorithm) could thus be repeated at this
stage. The outcome would be sets of clusters from the initially too finely
grained partition of sensorimotor states. As such, it seems like it might not
be necessary to introduce a further mechanism for estimating a suitable number
of clusters at each hierarchical level -- the method, when repeated across
levels of a hierarchy, could be seen as a hierarchical clustering method that
iteratively combines more and more clusters.

Another direction of future work is to explore how the presented method can be
utilized in the context of reinforcement learning. What the agent discovers is
a compact representation of latent states of the environment, which might be
used as feature representation. Indeed, it has been shown by Mahadevan and
Maggioni that a spectral transformation similar to the one used in this paper
can be used for the construction of a basis, to estimate a value function
\cite{mahadevan2007proto}. 

Finally, the role of the exploration policy in the discovery of sensorimotor
contexts, or contingencies, should be investigated. In this work, the agent
relied on a simple random walk as exploration policy, which, while sufficient,
certainly is a sub-optimal way to gather information about the sensorimotor
interaction and the environment. A more directed search strategy could
certainly speed up the learning process significantly.

\ifCLASSOPTIONcaptionsoff
  \newpage
\fi

\bibliographystyle{IEEEtran}
\bibliography{IEEEabrv,discovering_latent_state}

\begin{IEEEbiography}[{\includegraphics[width=1in,height=1.25in,clip,keepaspectratio]{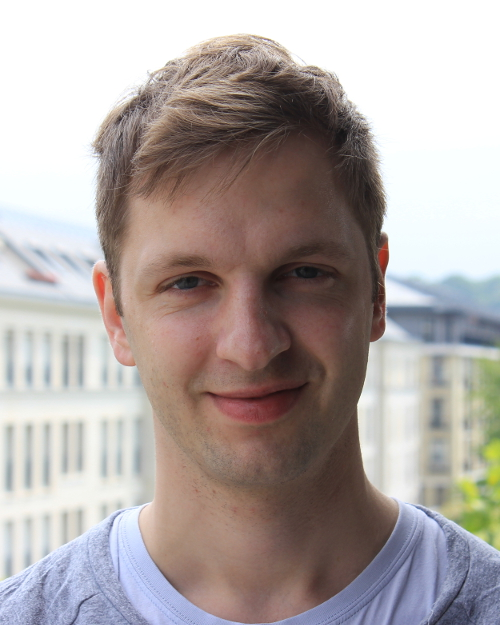}}]{Nikolas J. Hemion}

Dr. Nikolas Hemion received his Ph.D. degree in Intelligent Systems in 2013
from the Research Institute for Cognition and Robotics, Bielefeld University,
Germany. During his doctoral studies, he collaborated with the Honda Research
Institute Europe, and visited the Centre for Robotics and Neural Systems,
Plymouth University, UK. Subsequently he joined the AI Lab at Aldebaran as
senior researcher, and was appointed as director of the AI Lab in 2015.
Nikolas is also co-director of the H2020 innovative training network APRIL.
His research interests lie in cognitive architecture and model learning for
cognitive robotics, and self-organized learning of sensorimotor
representations.

\end{IEEEbiography}

\end{document}